\RequirePackage{tikz}
\documentclass[runningheads]{llncs}
\usepackage[T1]{fontenc}
\usepackage{graphicx}
\usepackage{lipsum}
\usepackage[absolute]{textpos}
\usepackage{afterpage}

\usepackage{url}
\usepackage{soul}
\usepackage{multirow}
\usepackage{setspace}
\usepackage{hyperref}

\usepackage{tikz}
\usetikzlibrary{shapes,positioning,fit}

\begin{document}
\title{Unsupervised Flow Discovery\\from Task-oriented~Dialogues}


%
%

\author{
Patrícia Ferreira\inst{1,2}
\and Daniel Martins\inst{3}
\and Ana Alves\inst{1,3}
\and Catarina Silva\inst{1,2}
\and Hugo Gonçalo~Oliveira\inst{1,2}
}
%
%
\institute{CISUC, Universidade de Coimbra, Portugal \and DEI, FCTUC, Universidade de Coimbra, Portugal \and ISEC, Instituto Politécnico de Coimbra, Portugal}
\maketitle              
%

\begin{abstract}

The design of dialogue flows is a critical but time-consuming task when developing task-oriented dialogue~(TOD) systems.
We propose an approach for the unsupervised discovery of flows from dialogue history, thus making the process applicable to any domain for which such an history is available.
Briefly, utterances are represented in a vector space and clustered according to their semantic similarity. Clusters, which can be seen as dialogue states, are then used as the vertices of a transition graph for representing the flows visually.
We present concrete examples of flows, discovered from MultiWOZ, a public TOD dataset.
We further elaborate on their significance and relevance for the underlying conversations and introduce an automatic validation metric for their assessment.
Experimental results demonstrate the potential of the proposed approach for extracting meaningful flows from task-oriented conversations.

\end{abstract}

\section{Introduction}

Dialogue systems are ubiquitous today.
Customer-support agents are used by companies to handle customer service inquiries and provide assistance; virtual assistants like Siri\footnote{\url{https://www.apple.com/siri/}} and Alexa\footnote{\url{https://alexa.amazon.com}} are  integrated into a range of smart devices to provide daily assistance through natural language interactions.

The aforementioned conversational agents are known as task-oriented systems, designed to fulfill distinct user tasks or requirements, such as making a reservation or requesting information. To cover all the target tasks, along with all possible interactions, creating dialogue flows for these systems is complex, time-consuming and often done manually. Furthermore, the resulting flows can rarely be shared across domains, meaning that their design has to be performed for each new dialogue system.

This is where the automatic discovery of dialogue flows can play a key role.
By analysing the history of conversational turns, this task may help in the identification of communication trends, which can support the design of flows for virtual agents, thus accelerating their processing, or help with their validation, possibly suggesting changes. 
Moreover, such flows enable to track the progression of the conversation over time and guide human agents, e.g.,~in a call centre, helping them to anticipate questions and provide prompt responses, without losing the benefits of human interaction~\cite{rapp2021human}.


 

We propose an approach for automatically discovering dialogue flows from a history of conversations.
It is unsupervised, thus applicable to any collection of task-oriented dialogues~(TODs), and follows three main steps.
Utterances are (i)~represented in a vector space and (ii)~clustered according to their semantic similarity;
(iii)~discovered clusters, which may be seen as dialogue states, are used as vertices of a transition graph.
For human consumption, the graph can be visually depicted, incorporating computed transitions and their corresponding probabilities, derived from the conversation history. For easier interpretation, states are labelled based on the most frequently occurring sequences within the clustered utterances.
This visual representation empowers users to conduct a post-inspection of the discovered flow, also useful for auditioning.

Towards its validation, the proposed approach is implemented and applied to a well-known dataset of TOD, MultiWOZ~\cite{Budzianowski2018}.
Resulting dialogue flows are depicted, which allows for a comprehensive understanding of the discovered patterns.
Furthermore, we propose an automatic measure for computing to what extent the transitions in the test portion of the dataset align with the flows discovered from the training portion. This establishes the reliability and accuracy of our approach.



The remainder of the paper is structured as follows:
Section~\ref{sec:related} reviews previous work related to dialogue flow discovery;
Section~\ref{sec:approach} describes the proposed approach;
Section~\ref{sec:experimentation} describes how 
it was experimented, including the used dataset and implementation details;
Section~\ref{sec:results} presents the resulting flows, followed by an attempt to their automatic evaluation;
Section~\ref{sec:conclusion} concludes the paper and provides cues for future work.

\section{Related Work}\label{sec:related}

Utterances in dialogues are commonly classified according to: (i) user intents, e.g., make a reservation, find attractions; or (ii) dialogue acts, more generic, relative to the action performed by the speaker, e.g., ask, explain, request.


To some extent, the significance of the previous labels can be overlooked by data-driven chatbots, such as sequence-to-sequence agents~\cite{VinyalsL15}, also including agents built on top of large language models, e.g., ChatGPT\footnote{\url{https://chat.openai.com/}} or Bard\footnote{\url{https://bard.google.com/}}.
However, such labels are highly relevant for task-oriented systems, which rely on the dialogue structure to achieve specific goals efficiently. Their incorporation facilitates successful interactions and goal completion, setting them apart from more general conversational models.

The automatic classification of utterances is typically done in a supervised fashion~\cite{cohen2004learning,firdaus2021deep}, thus requiring annotated data, which may not be readily accessible in numerous domains. Hence, the development of an approach that does not rely on annotated data has the potential to significantly expand the range of application scenarios.

The design of dialogue flows in task-oriented systems can steer the conversation in specific directions, avoiding purely reactive responses~\cite{grassi2022knowledge}.
It encompasses the definition of task-specific intents and of training phrases, among other decisions, and generally ends up being created manually, often with the help of tools like Google's DialogFlow\footnote{\url{https://cloud.google.com/dialogflow/}}, Microsoft Luis\footnote{\url{https://www.luis.ai/}}, or Rasa\footnote{\url{https://rasa.com/}}.

To automate this process, it is necessary to cluster expressions that represent the same intent, the same action, or mention the same information.
For this, utterances are typically represented in a vector space where those semantically similar are closely positioned together. Methods like word~\cite{hashemi2016query} or sentence~\cite{liu2021open,park2022analysis} embedding serve this purpose, enabling to group utterances with comparable meanings or content.



Towards intent discovery, utterances may be clustered with well-established algorithms like k-means, directly applied to the utterance embeddings~\cite{park2022analysis,liu2021open}, possibly incorporating additional contextual features~\cite{zhang2021discovering}.



Some related works represent dialogue flows as graphs, with the discovered clusters as the vertices~\cite{joty2011unsupervised,bouraoui2017cluster,ritter2010unsupervised}. An earlier approach~\cite{ritter2010unsupervised} tackles the same problem, employing a Hidden Markov Model~(HMM) when learning from Twitter conversations.
Their approach introduces interesting features, such as the vertices for representing the start and end of dialogue, and a threshold for ignoring transitions with low probability. 
Clusters were labelled manually, while our work aims to streamline and enhance the efficiency of dialogue flow discovery, also reducing the manual efforts for cluster~labelling.

Clusters group similar utterances without considering the sequence of conversation~\cite{joty2011unsupervised}.
But graphs may represent the main transitions between topics and dialogues~\cite{bouraoui2017cluster}. Different methods were proposed to model such transitions, e.g., HMM~\cite{ritter2010unsupervised}, Latent Dirichlet Allocation \cite{li2015supervised} or Deep Learning techniques~\cite{serradilla2022deep}.


Graph2Bots~\cite{bouraoui2019graph2bots} is a tool for assisting agent conversation designers in extracting graphs of human-to-human conversations. The resulting graph includes the main dialogue phases and their transitions.
An alternative approach finds the most frequent sequences of questions and responses when representing their sequences by a finite-state automaton~\cite{sastre-martinez-nugent-2022-inferring}.
Despite relying on intent labels and on proprietary software, the previous work enables to visualize the flows and was demonstrated for the restaurant booking dialogues in the MultiWOZ~dataset.
None of the previous undergoes a quantitative evaluation of the resulting flows.




We propose both a new approach for flow discovery and a metric for flow validation in unseen dialogues.
Given a history of dialogues, similar utterances are clustered;
a graph is created based on cluster transitions;
and a visualisation is provided, 
to support TOD systems and their~users.

\section{Unsupervised Dialogue Flow Discovery}\label{sec:approach}

\begin{figure}[!ht]%
\centering
\input{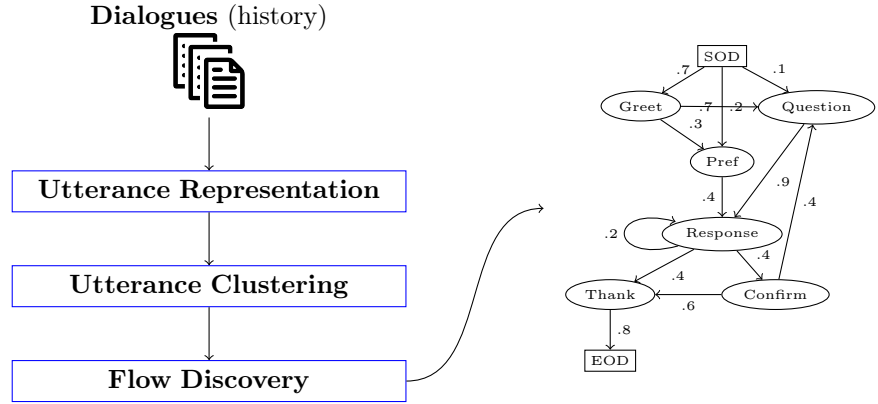}
\caption{Overview of the proposed approach with three steps: utterance representation, utterance clustering, and flow discovery.}\label{fig:approach}
\end{figure}

We propose a generic approach for automatically discovering the most frequent flows in a history of dialogues, and for representing them visually, as a transition graph.
Figure~\ref{fig:approach} summarises this approach, which encompasses three steps: utterance representation, utterance clustering, and flow discovery.

In order to apply our approach to a specific domain, a history of dialogues covering all the relevant intents for that domain is required. Dialogues are represented as written sequences of utterances contributed by various participants. While our approach is versatile and can be adapted to various scenarios, our primary focus is on TOD involving a customer and an agent.

In Figure~\ref{fig:approach}, from top to bottom, in the first step, the dialogues are represented according to their meaning, for example, through their embeddings.
In the second step, the dialogues are grouped based on their semantic similarity, with the intention that these groupings represent dialogue states.
In the third step, transitions between the states and their corresponding probabilities are calculated. This results in a transition graph $G(C, T)$, with the states $c \in C$ as vertices and transitions $t(c_a, c_b, p_{ab}) \in T$ as edges, where $p_{ab}$ is the probability of transitioning from state $c_a$ to state $c_b$. The latter is calculated according to Equation~(\ref{eq:prob}), where $|t(c_a,c_b)|$ represents the number of dialogues in $c_b$ that immediately follow a dialogue in $c_a$.



\begin{equation}\label{eq:prob}
    p_{ab} = \frac{\| t(c_a,c_b)\| }{\sum_{x \in C} \|t(c_a,c_x\|}
\end{equation}

In addition to the vertices derived from the clusters, two vertices representing the Start of Dialogue (\texttt{SOD}) and the End of Dialogue (\texttt{EOD}) are included.
Moreover, to ease human interpretation, states can be identified by automatically-assigned human-labels. These can be based on frequent sequences or keywords used in the clustered utterances.



\section{Experimentation Setup} \label{sec:experimentation}

This section describes the experimentation designed for testing the proposed approach, focused on its application to a dataset of TOD. 
After described this dataset, implementation details are provided.

\subsection{Dataset}\label{sec:dataset}

Experimentation was conducted with MultiWOZ 2.2~\cite{zang-etal-2020-multiwoz}, a well-known dataset of TOD.
Its dialogues include multiple turns of interactions between two humans: a USER plays the role of a user, who has a task to fulfill; a SYSTEM tries to answer requests promptly, while helping the user to conclude the task.
Dialogues cover a total of eight domains~(restaurant, attraction, hotel, taxi, train, bus, hospital, police) and several intents, such as finding a restaurant or booking a taxi.

Besides the domain, other annotations~(e.g., intent, slots) are available, but none of them are used in this work.
This kind of information is rarely available, and would have to be predicted by an automatic classifier.
Moreover, since intents change according to the domains of the dialogue, considering them would make our approach no longer domain-agnostic.


MultiWOZ~2.2 is divided into train and test portions, distributed according to Table~\ref{tab:Multiwozdistribution}.
Dialogues are sequences of turns, each including a USER utterance and a SYSTEM response.
In this work, flows are discovered from the train portion, with the test portion held out for validation.
Table~\ref{tab:dialogues} illustrates the contents of the dataset with a single dialogue.


\begin{table}[!ht]
\caption{Contents of MultiWOZ~2.2 dataset.}\label{tab:Multiwozdistribution}
\centering
\begin{tabular}{r|r|rrr}
\hline
\multirow{2}{*}{\textbf{Portion}} & \multirow{2}{*}{\textbf{\#Dialogues}} & \multicolumn{3}{c}{\textbf{\#Utterances}} \\
&  & \textbf{Total} & \textbf{USER} & \textbf{SYSTEM}\\
\hline
Train & 8,436                    & 113,552       &  56,776  &      56,776         \\
Test  & 1,000                    & 14,744   &     7,732     &    7,732     \\
\hline   
\end{tabular}%
\newline
\end{table}

\begin{table}[!ht]
\centering
\footnotesize
\caption{Dialogue SNG1066 from the MultiWOZ~2.2 test portion.}
\label{tab:dialogues}
\begin{tabular}{lp{0.88\textwidth}}
\hline
\textbf{Speaker} & \textbf{Utterance}                                                                                                                          \\ \hline
USER             & Could you help me find a boat to visit on the north side?                                                                                    \\
SYSTEM           & I have one in that area. It's called the Riverboat Georgina. It's located at Cambridge Passenger Cruisers, Jubilee House. Would you like their phone number for more information?                                    \\
USER             & Yes, I want the phone number and also the entrance fee, please.                                                                                       \\
SYSTEM           & Their phone number is 01223902091 and we do not have the entrance fee in our database at this time.                                                \\
USER             & Okay, that is all I need today. Thank you very much.                                            \\
SYSTEM           & You're very welcome! Thanks for contacting the Cambridge TownInfo Centre and have a great day! \\
 \hline      
\end{tabular}
\vspace{-.5cm}
\end{table}

\subsection{Implementation}\label{sec:implementation}


Different methods were tested for utterance representation, clustering and labelling.
Yet, except for the latter, we focus on a single implementation and leave the comparison of alternatives for future work.

As long as it allows to measure the semantic similarity between different utterances, further enabling similarity-based clustering, any vector representation can be adopted for the utterances.
We used the 384-sized embeddings obtained directly from the model \texttt{all-MiniLM-L6-v2}, available from the Sentence Transformers library\footnote{\url{https://www.sbert.net/}}.
This is reported as one of the best-performing models for sentence embedding and also one of the smallest~(80MB).

Any clustering algorithm could have been used in the second step, but we chose the k-means algorithm, available in the scikit-learn library\footnote{\url{https://scikit-learn.org/stable/modules/generated/sklearn.cluster.KMeans.html}}, due to its widespread use in clustering tasks.
k-means was used with a maximum of 1,000 iterations and centroids initialized by sampling, based on an empirical probability distribution of points~(\texttt{k-means++}).
For MultiWOZ 2.2, the number of clusters, denoted as $k$, was determined with Silhouette method, which assesses the cohesion and separation of clusters.

\begin{figure}
\vspace{-.6em}
\centering
\includegraphics[width=.6\textwidth]{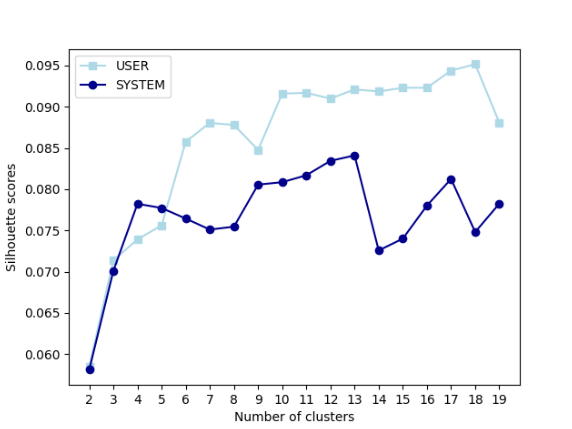}
\caption{Silhouette scores for different values of $k$ for USER and SYSTEM utterances.} \label{fig:silhouette}
\vspace{-.6em}
\end{figure}

As we always know who the speaker of each utterance is, user and system utterances were analyzed separately, resulting in a different number of clusters for each. A scatter plot of Silhouette scores for various values of $k$ is illustrated in Figure~\ref{fig:silhouette}. The maximum score is when $k=18$ and $k=13$, respectively for the user and the system. These were the numbers of clusters used.

On the flow discovery, besides computing the transitions and their probabilities as in Section~\ref{sec:approach}, two approaches were tested for labelling the clusters.
One uses the most frequent verb phrases~(VPs) in their utterances, obtained from the model \texttt{en\_core\_web\_md} of the spaCy\footnote{\url{https://spacy.io/}} toolkit;
the other relies on KeyBERT~\cite{grootendorst2020keybert} for extracting the most frequent keywords in the utterances of the cluster.

\section{Results}\label{sec:results}

This section summarizes the results obtained with the previously described implementation of the approach.
It presents resulting dialogue flows and describes an attempt at the automatic evaluation of their effectiveness. 

\subsection{Discovered Flows}

Following the procedure described in Section~\ref{sec:implementation}, utterance clustering resulted in 18 USER and 13 SYSTEM clusters~(i.e., states).
These could be identified by a number and simply listed, but this would not provide enough information on the dialogue state.

However, dialogue states, alone, are not enough for representing the flows, which should include transitions between states.
As mentioned in Section~\ref{sec:approach}, dialogue flows can be represented by graphs, but the ideal visualisation is not straightforward.
Besides the challenge of generating informative labels, a graph with all the states and transitions will contain some rare transitions and can be too complex to manage.
On the other hand, a graph with few states will be easier to read at the cost of losing coverage of a fraction of transitions.
Therefore, we look at graphs after the application of different thresholds $\theta$ on the transitions.

\begin{figure}[]
\centering
\includegraphics[width=1.15\linewidth]{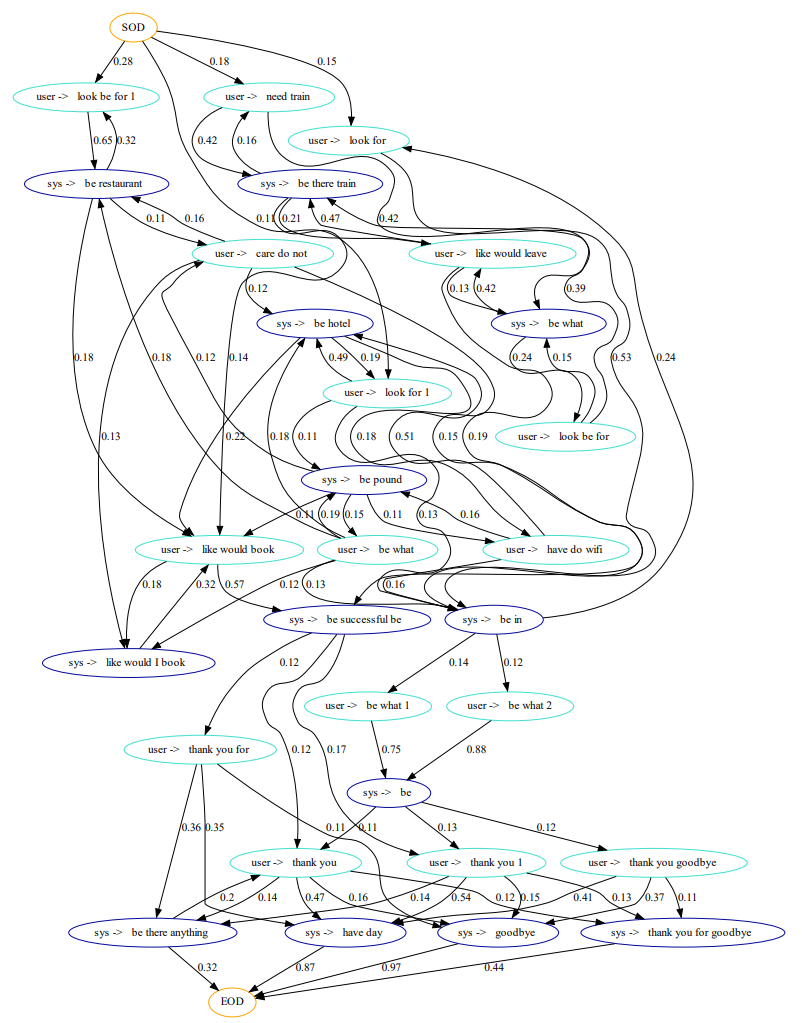}
\caption{Flow discovered from the MultiWOZ train portion with $\theta=0.10$ and labels generated from the most frequent verb phrases.} \label{fig:treshold10}
\end{figure}

\begin{figure}[h!]
    \centering
\includegraphics[width=0.6\linewidth]
{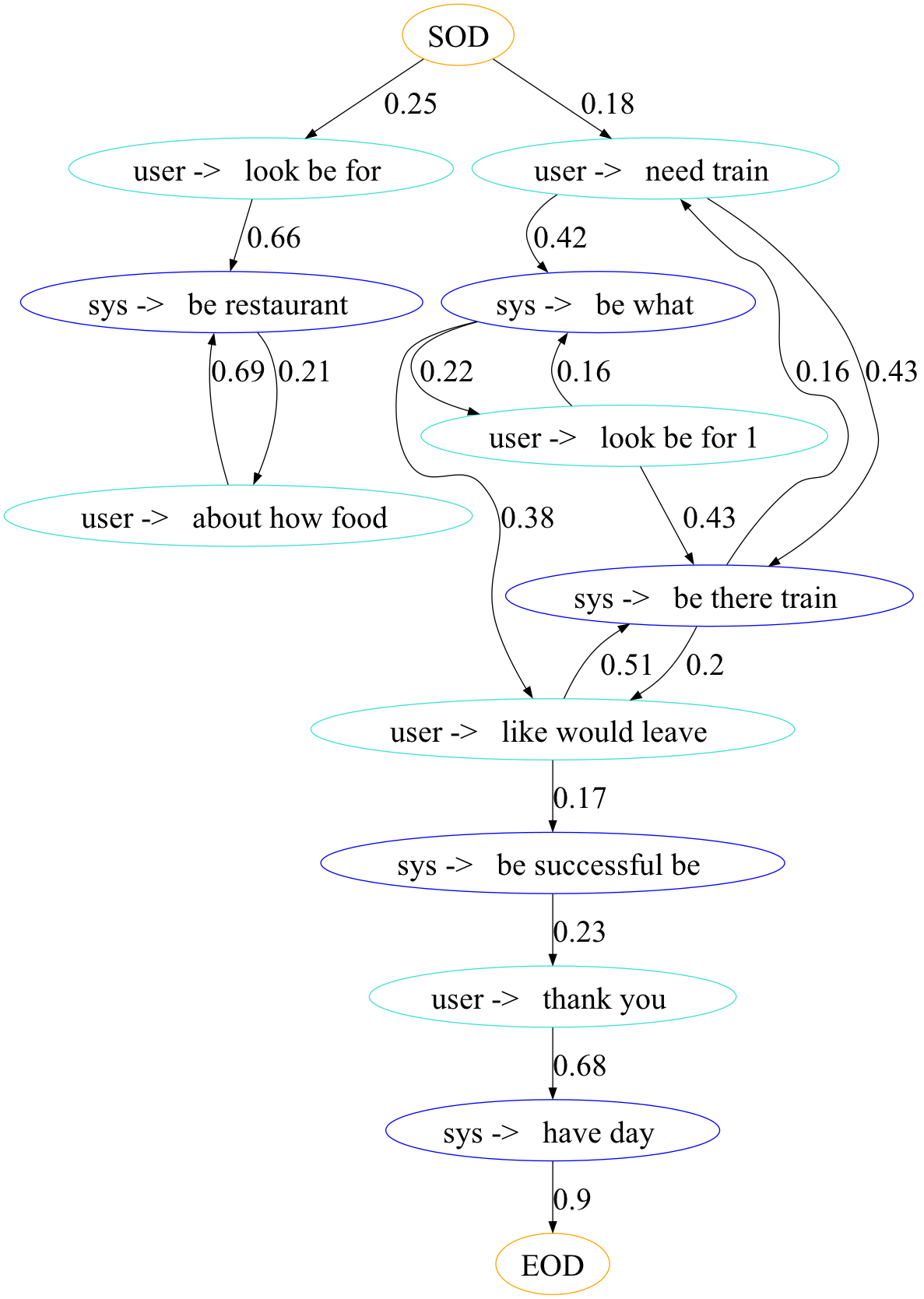}
    \caption{Flow discovered from the MultiWOZ train portion with $\theta=0.15$ and labels generated from the most frequent verb phrases.}
    \label{fig:treshold15}
  \vspace{-.3cm}
  \end{figure}

Figures~\ref{fig:treshold10} and~\ref{fig:treshold15} depict the discovered flows, respectively for $\theta=0.10$ and $\theta=0.15$.
Vertices in light blue represent the USER; in dark blue represent the SYSTEM; and in yellow correspond to the start~(\texttt{SOD}) and end~(\texttt{EOD}) of dialogue.

Labels give insight on the system's actions and responses, such as hotel suggestions, available options or greetings, and reveal user's needs and intentions, such as hotel search, help and specific information. If different clusters end up with the same label, a unique number is appended.
The VP method tends to result in grammatically well-constructed labels, whereas KeyBERT relies on the most relevant terms.

By increasing the value of $\theta$, lower-probability transitions are ignored, resulting in more simplified graphs focused on the strongest relationships between clusters. Conversely, by reducing the value of $\theta$, less frequent transitions can be included in the visualisation, providing more detailed insights into the possible interactions that may occur during the dialogue.

When analysing these flows, we notice 
common patterns in the dialogues of MultiWOZ.
They usually begin with a request or a question by the user, related to their specific need, but generally without an isolated greeting, common in social interactions. This might be due to the task-oriented nature of the dataset, where users have clear goals and seek prompt answers to their needs.

As the dialogue progresses, the interaction between the user and the system unfolds, with exchanges of information, answers and clarifications to meet the user's needs. This intermediate phase may include several steps, such as the system providing options or additional details, and the user making choices or asking for more specific information.

Finally, upon reaching the end of the dialogue, it is common to find a conclusion or fulfilment step, where the system may thank the user for the interaction, provide final information, or say goodbye.

\subsection{Automatic Evaluation}

Flow visualizations provide interesting insights, but drawing conclusions on their utility and on how well they represent the underlying dialogues is challenging, especially without experts in the dataset's domains and call-center procedures. Yet, quantitatively evaluating flows is complex and lacking in related work~\cite{bouraoui2019graph2bots,sastre-martinez-nugent-2022-inferring}.

Towards a quantitative metric of success, we propose to assess the flow by measuring to what extent an unseen portion of dialogues~(i.e., test set) follows the flows discovered from another portion of the same type (i.e., train set).
Briefly, this evaluation encompasses the following steps:
\begin{enumerate}
    \item Discover flows from dialogues in the train set.
    \item For each utterance $u_x$ in the test set:
    \begin{enumerate}
        \item assign $u_x$ to the most similar cluster in the flow, $c_x \in C$;
        \item get the utterance that directly follows $u_x$, $u_y$;
        \item assign $u_y$ to the most similar cluster in the flow, $c_y \in C$;
        \item check whether the transition $t(c_x, c_y, p_{xy})$ is in the flow~(i.e, is predicted).
        
    \end{enumerate}
\end{enumerate}

For different values of $\theta$, we calculated the \textbf{transition accuracy}, i.e., the proportion of predicted transitions in relation to all the transitions in the test set. 
Obtained values are in Table~\ref{tab:eval}.

The higher $\theta$ is, the smaller the graph, and the lower the accuracy.
With $\theta=0.20$, there are no paths between \texttt{SOD} and \texttt{EOD}, so no transitions are predicted.
With $\theta=0.05$, more than 80\% of the transitions in the test portion are predicted.
This is promising, especially considering that it is based on unseen dialogues, with previously unseen utterances, where language may vary.
Nevertheless, we should note that requests and responses in MultiWOZ tend to be direct.
And still, we see a significant number of transitions~($\approx18\%$) that follow rare paths.

\begin{table}[!ht]
    \centering
    \setlength{\tabcolsep}{8pt}
    \caption{Configuration of the dialogue flows discovered from the MultiWOZ~2.2 train portion, with different values of threshold $\theta$; and transitions accuracy in the MultiWOZ~2.2 test portion, as computed by the previous flow.}
    \label{tab:eval}
    \begin{tabular}{c c c c }
        \hline
        $\theta$ & $|C|$ & $|T|$ & \textbf{Accuracy} \\
        \hline
        0.05 & 33 & 139 & 82.2\% \\
        0.10 & 32 & 66 & 70.4\% \\
        0.15 & 23 & 32 & 43.6\% \\
        0.20 & 0 & 0 & 0 \\
        \hline
    \end{tabular}
\vspace{-1em}
\end{table}



This score is not comparable to an evaluation by human experts, but it is a fast way of obtaining initial quality-related figures.
Moreover, it enables the comparison of flows discovered with different parameters and approaches.
In the future, we will attempt to increase accuracy, by testing different embeddings or clustering algorithms, ideally without increasing the number of clusters.


%
\section{Conclusion and Future Work}\label{sec:conclusion}
Motivated by the importance of flows in TOD systems, and by the amount of manual work involved in their production, validation, and maintenance, we presented an innovative approach for unsupervised flow discovery.
To provide a comprehensive analysis of communication trends, it follows a three-step method that: represents utterances as vectors; clusters semantically similar ones to form dialogue states; and constructs a transition graph with such states as vertices.

Experiments showcase the potential of our approach, providing a pathway for extracting meaningful dialogue flows without the need for annotated data. The graphical visualisations offer an intuitive representation, supporting both developers and users in interpreting and optimising TOD systems.
This is complemented with the proposal of an automatic evaluation, which shows that the flows represented by 33 dialogue states can predict more than 80\% of the transitions in the test portion of MultiWOZ~2.2, a widely-used dataset in the development of TOD systems.

There is room for improvement, but progress can now be measured with the proposed metric.
For clustering, it would be interesting to: explore other algorithms, including density-based, which do not require the number of clusters as input; consider the context of previous utterances as contextual features; extensively assess the impact of different models for utterance embedding.
Another line would be investigating different methods for labelling clusters and visualising dialogue flows, towards deeper insights and more informative representations.
The approach should be further validated in diverse data, including less artificial dialogues. We are currently working with industry partners and have plans to soon apply it to real call-center data.
This should provide a more comprehensive evaluation of its effectiveness, and may be further complemented by a human-expert evaluation, hopefully offering valuable assessments and insights.

Towards more trustworthy machines, it would also be interesting to discover flows from virtual agents that are not flow-based, such as data-driven chatbots~(e.g., ChatGPT and Bard), thus contributing to their explanation.






\subsubsection{Acknowledgements} 
{\footnotesize This work was partially supported by: the project
FLOWANCE (POCI-01-0247-FEDER-047022), co-financed by the European Regional Development Fund, through PT2020, and by the Competitiveness and Internationalization Operational Programme;
the Portuguese Recovery and Resilience Plan through project C645008882-00000055, Center for Responsible AI;
and by national funds through FCT, within the scope of the project CISUC (UID/CEC/00326/2020).}

%
%
%
\begin{spacing}{.95}
\bibliographystyle{splncs04}
\bibliography{bibliography}    
\end{spacing}
%




\end{document}